\documentclass[11pt,a4paper]{article}
\usepackage{hyperref}
\usepackage[nohyperref]{emnlp2020}
\usepackage{times}
\usepackage{tikz}
\usepackage{latexsym}
\usepackage{graphicx}
\usetikzlibrary{shapes}
\usetikzlibrary{positioning}
\usetikzlibrary{arrows,automata}

\usepackage{linguex}

\usepackage{microtype}

\aclfinalcopy 

\title{Investigating Novel Verb Learning in BERT: Selectional Preference Classes and Alternation-Based Syntactic Generalization}

\author{Tristan Thrush $^1$, Ethan Wilcox $^2$, and Roger Levy $^1$ \\
  $^1$ MIT Department of Brain and Cognitive Sciences / 43 Vassar St, Cambridge, MA, 02139, USA\\
  $^2$ Harvard Department of Linguistics / Boylston Hall, 3rd floor, Cambridge, MA 02138, USA\\
  \texttt{tristant@mit.edu, wilcoxeg@g.harvard.edu, rplevy@mit.edu}\\}

\date{}

\begin{document}
\maketitle
\begin{abstract}

Previous studies investigating the syntactic abilities of deep learning models have not targeted the relationship between the strength of the grammatical generalization and the amount of evidence to which the model is exposed during training. We address this issue by deploying a novel word-learning paradigm to test BERT's \cite{devlin2018bert} few-shot learning capabilities for two aspects of English verbs: alternations and classes of selectional preferences. For the former, we fine-tune BERT on a single frame in a verbal-alternation pair and ask whether the model expects the novel verb to occur in its sister frame. For the latter, we fine-tune BERT on an incomplete selectional network of verbal objects and ask whether it expects unattested but plausible verb/object pairs. We find that BERT makes robust grammatical generalizations after just one or two instances of a novel word in fine-tuning. For the verbal alternation tests, we find that the model displays behavior that is consistent with a transitivity bias: verbs seen few times are expected to take direct objects, but verbs seen with direct objects are not expected to occur intransitively. The code for our experiments is available at \url{https://github.com/TristanThrush/few-shot-lm-learning}.

\end{abstract}

\section{Introduction}

Contemporary deep learning models for language have been shown to learn many aspects of natural language syntax including a number of long-distance dependencies \cite{gulordava2018colorless, marvin2018targeted, wilcox2018rnn}, selectional properties of verbs \cite{kann2019verb}, representations of incremental syntactic state \cite{futrell2019neural} and information from which hierarchical structure can be linearly decoded \cite{hupkes2018visualisation, hewitt2019structural,lakretz2019emergence}.  These and many other related studies demonstrate an impressive range of human-like linguistic knowledge that is automatically acquired by these models simply from exposure to large quantities of raw text.  However, human-like grammatical abilities include not just rich and detailed linguistic knowledge but the ability to deploy this knowledge in using new words based on minimal exposure \citep{carey-bartlett:1978acquiring,gropen1989learnability,perek2017linguistic}.  It remains poorly understood what grammatical generalizations contemporary deep learning models are able to make regarding the behavior of words to which they have minimal exposure. In this work, we assess the syntactic generalization behavior of a contemporary neural network model (BERT; \citet{devlin2018bert}) on two novel phenomena in English  and address the question of single-shot and few-shot learning, demonstrating that BERT makes robust grammatical generalizations after fine-tuning on minimal examples of a novel token.

We test BERT's few-shot learning capabilities on two phenomena at the syntax-semantics interface: English verbal alternations, and verb/object selectional preferences. In English, verbs can appear in multiple syntactic frames; which frame a verb appears in is governed by its argument structure properties. Often, frames are paired into alternation classes \cite{levin1993english} such that when English speakers hear a novel verb in one frame they can be confident that it can be used in its alternation-class pair. Using the well-attested \textit{dative alternation} as an example, if a listener hears the sentence ``I \textit{daxed} the tennis racket to my friend" they would expect that ``I \textit{daxed} my friend the tennis racket" is a grammatical English sentence, meaning approximately the same thing. They would not, however, have such an expectation for ``I \textit{daxed} my friend for the tennis racket." In addition, listeners may be attuned to semantic clustering of verbal arguments based on past experience. For instance, following the example above, English speakers may expect \textit{dax} to take an animate indirect object, and would find examples such as ``I \textit{daxed} the court the tennis racket" to be surprising. 

We take inspiration for our testing regime from a class of psycholinguistic experiments known as `novel word learning studies', which we adapt to the neural setting. In such experiments subjects are exposed to a novel word in context during a training phase, and assessed for what grammatical generalizations they have learned about the novel word during a later testing phase. Novel word learning experiments have been used to assess human grammatical generalization since \citet{berko1958child}, and have been deployed to assess semantic, as well as syntactic, generalizations \cite{carey-bartlett:1978acquiring}.
In this work, we replicate the novel word learning paradigm in the neural setting by fine-tuning BERT on tightly-controlled sentences that contain novel verbs and objects, and assessing the model on carefully constructed test sets that reveal what grammatical generalizations it has learned. We find that BERT is able to make proper generalizations for both verbal alternations as well as semantic clustering for verbal arguments after just one or two exposures during training.

\section{Methods}
\label{sec:methods}

For each test, we fine-tune BERT with sentences that contain new tokens for novel words. We then assess the the model's learning outcomes in one of two testing settings, described below.\footnote{For detailed information model architecture and training, see Appendix B. Unless otherwise noted, statistical tests are the result of linear mixed effects models with maximal random effects structure as advocated in \cite{barr2013random}.}

\subsection{Fine-Tuning}
We fine-tune BERT with its masked-language modeling objective to predict each of the novel verb tokens in the training data.  We add a new output neuron in the language modeling head, and a new embedding, for each novel word. In order for exposure during fine-tuning to approximate the effect of exposure to low-frequency words during the initial training, we optimize only newly-added weights.

During fine-tuning we mask all open-class content words that are not targeted by the experiment, and add determiners if they can be useful at designating the category of a masked word. Sample fine-tuning sentences are given in \ref{ex:fine-tuning-alternation} for our alternation tests and \ref{ex:fine-tuning-verbclasses} for our verb selectional preference tests.

\ex. 
\a. The [MASK] will [\textit{dax}] the [MASK] to the [MASK] \label{ex:fine-tuning-alternation}
\b. The [MASK] [\textit{daxed}] the [\textit{blicket}] \label{ex:fine-tuning-verbclasses}

Masking content words means that the model must rely on purely syntactic information such as word-order, prepositions and auxiliary verbs for its syntactic generalizations. We also control for tense within our experiments by using the same verbal tense across conditions within a training context.

\subsection{Evaluation}

\paragraph{Psycholinguistic Generalization Test:} Following \citet{linzen2016assessing} and \citet{futrell2018rnns}, we gauge BERT's learning outcomes by deriving the novel verb's probability in paired contexts in which the novel token's use is consistent with the training data plus grammatical rules (the \textit{in-class} context) or inconsistent with the data and the rules (the \textit{out-class} context). If the token is more likely in the \textit{in-class} context, then the model can be said to have learned the proper syntactic generalization. For these tests we report the proportion of the time the token is more likely in the \textit{in-class} contexts across 200 randomly-seeded training runs. The probability of a token, [T], is derived in the standard way from BERT by inserting a [MASK] token in it's place, and taking BERT's contextualized word embedding of this [MASK] token. This embedding is fed into BERT's language modelling head, which returns a probability for the token, [T], given the context.

\paragraph{Embedding Classification Test:} We also test BERT by probing the learned representations of embeddings for novel verb tokens directly (we use this method only for the alternation tests). In this testing procedure we train a linear model to predict whether a pre-trained BERT embedding corresponds to a verb that is in a particular alternation class, for example whether it follows the dative alternation or not. We then use the classifier to predict whether the novel verb is a member of the alternation class. Our linear classifiers achieve a mean accuracy of 0.992 on their training set. For the test set, we also report accuracy scores across 200 model runs.

\section{Selectional Preferences}

\begin{figure*}[h!]
\centering
\begin{tikzpicture}
\node[rectangle, fill=cyan!30] at (-.2,0) {Verb6};
\node[ellipse, fill=cyan!30] at (3.5,0) {Noun6};
\node[rectangle, fill=cyan!30] at (-.2,1) {Verb5};
\node[ellipse, fill=cyan!30] at (3.5,1) {Noun5};
\node[rectangle, fill=cyan!30] at (-.2,2) {Verb4};
\node[ellipse, fill=cyan!30] at (3.5,2) {Noun4};
\node[rectangle, fill=green!30] at (-.2,3) {Verb3};
\node[ellipse, fill=green!30] at (3.5,3) {Noun3};
\node[rectangle, fill=green!30] at (-.2,4) {Verb2};
\node[ellipse, fill=green!30] at (3.5,4) {Noun2};
\node[rectangle, fill=green!30] at (-.2,5) {Verb1};
\node[ellipse, fill=green!30] at (3.5,5) {Noun1};
\draw[thick]

(0.5,0) -- (2.5,0)
(0.5,0) -- (2.5,1)
(0.5,1) -- (2.5,0)
(0.5,1) -- (2.5,2)
(0.5,2) -- (2.5,2)
(0.5,2) -- (2.5,1)

(0.5,3) -- (2.5,3)
(0.5,3) -- (2.5,4)
(0.5,4) -- (2.5,3)
(0.5,4) -- (2.5,5)
(0.5,5) -- (2.5,5)
(0.5,5) -- (2.5,4)
;

\draw[dashed]
(0.5,0) -- (2.5,2)
(0.5,1) -- (2.5,1)
(0.5,2) -- (2.5,0)

(0.5,3) -- (2.5,5)
(0.5,4) -- (2.5,4)
(0.5,5) -- (2.5,3)
;
\end{tikzpicture}
\includegraphics[width=0.3\textwidth]{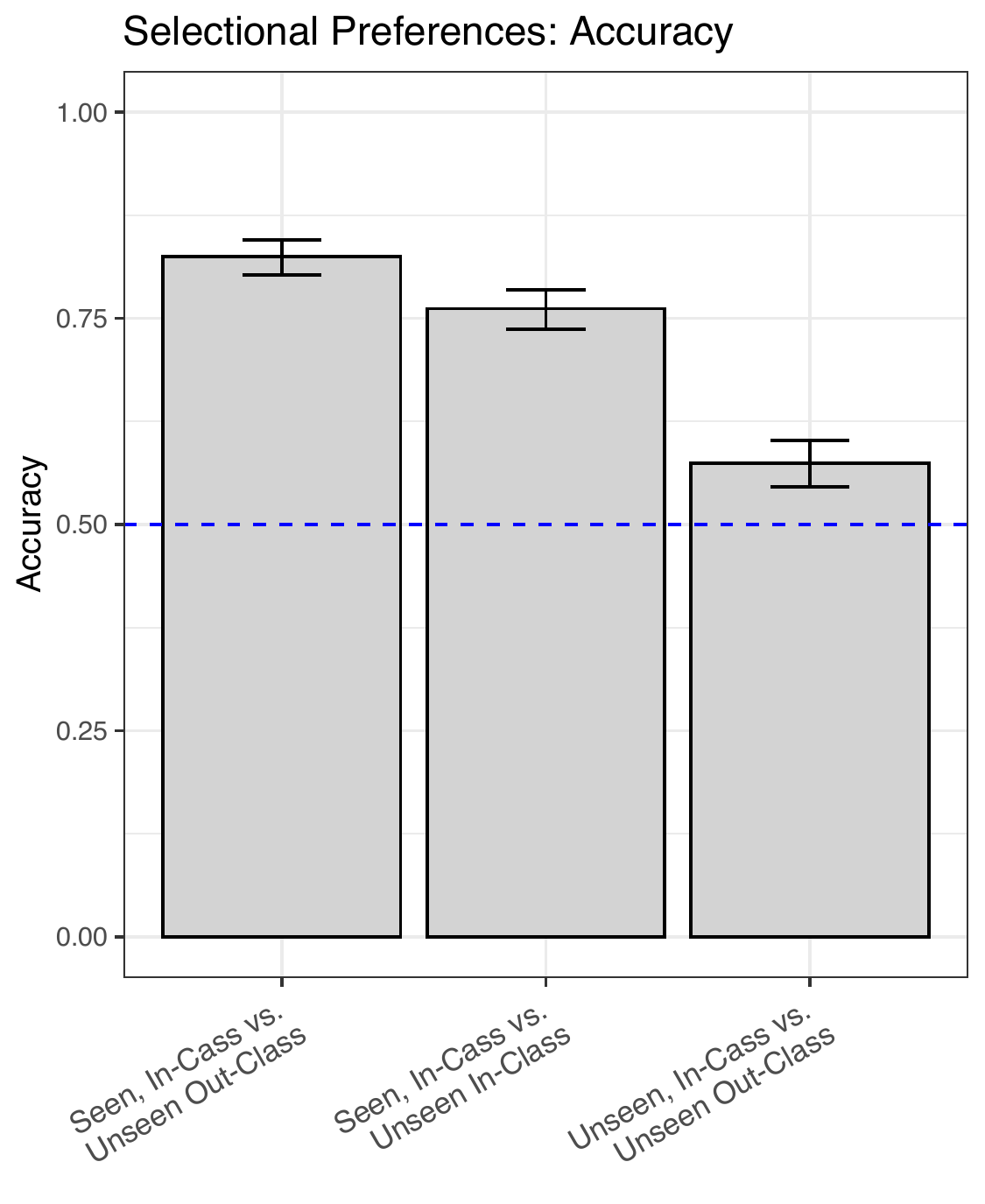}
\includegraphics[width=0.3\textwidth]{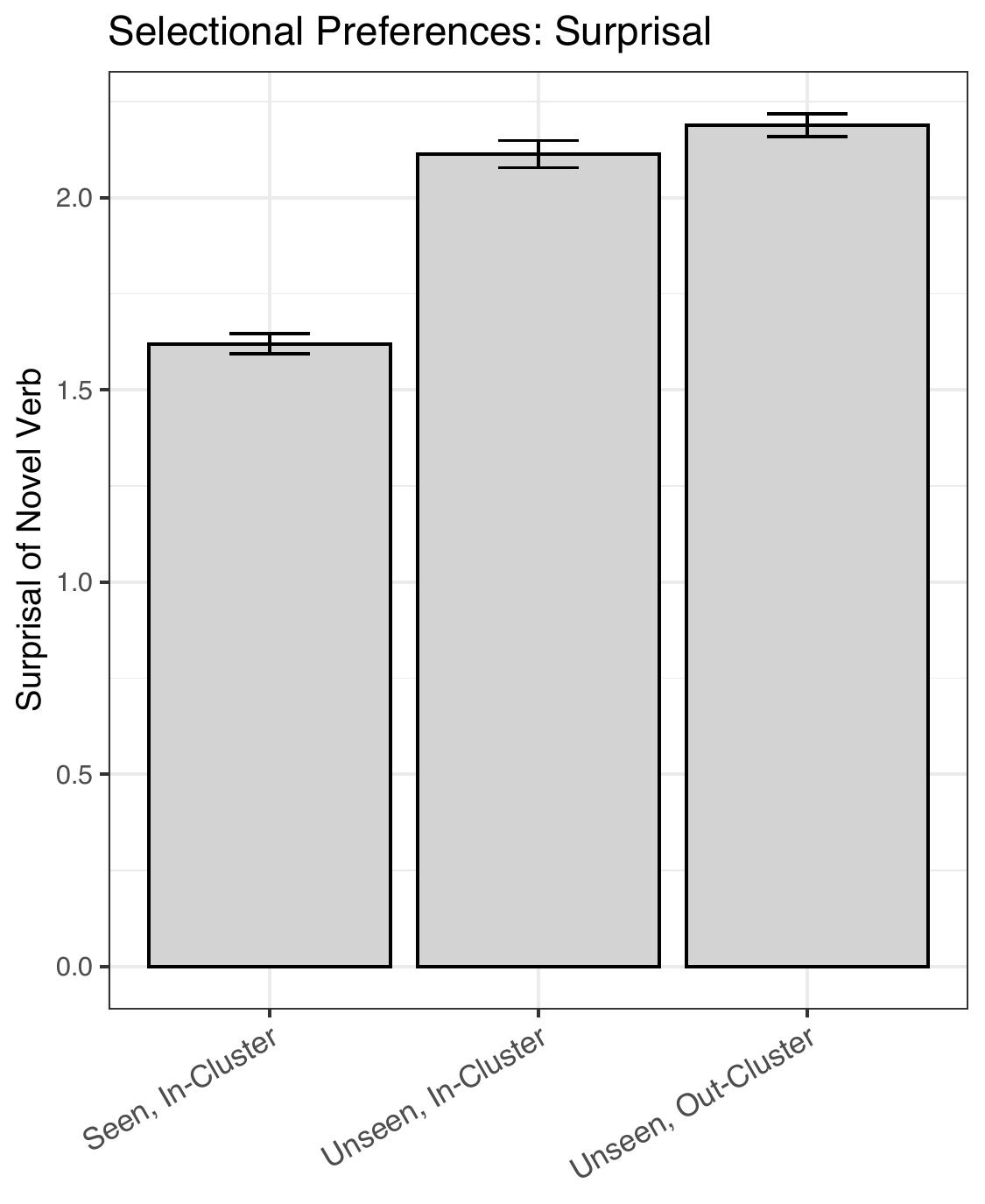}\\
\makebox[\columnwidth][s]{(a) (b) (c)}
\caption{(a): Selectional restrictions imposed on the 6 nonce verbs and 6 nonce nouns in the fine-tuning data. Each verb (rectangled) appears with two nouns (circled), such that the full selectional paradigm for the verb must be inferred. (b) and (c): Results from our selectional preference tests, showing significant difference between all contrasts tested.}
\label{fig:selectional-restrictions}
\end{figure*}
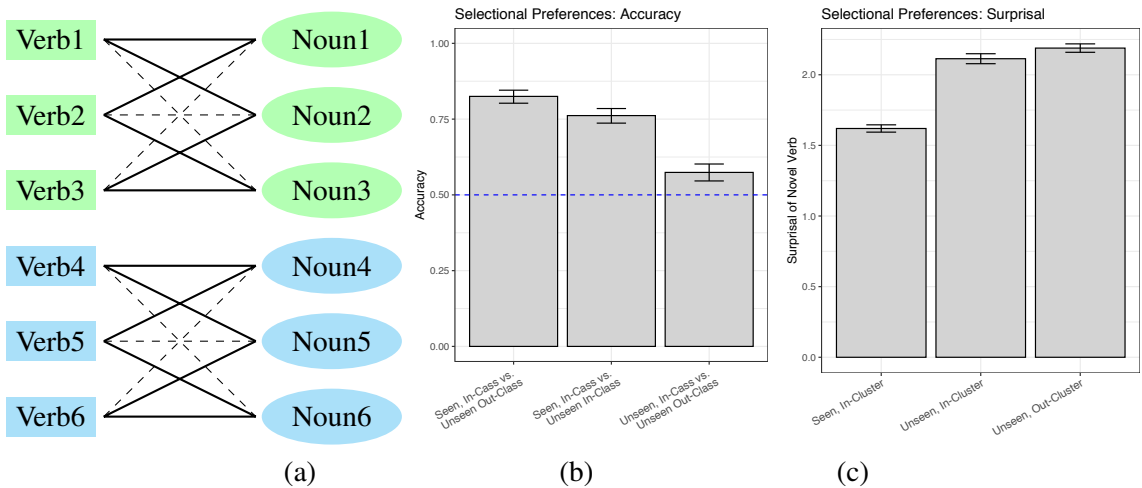

Verbs can impose a variety of selectional restrictions on semantic properties of nouns to limit which clusters of nouns they accept. Just to name a few, these restrictions can require an object to be animate or inanimate, a location, or a raw material \cite{levin1993english}. In this section, we ask what generalizations BERT makes about a verb's selectional restrictions based on incomplete, limited exposure. For our experiments, we define selectional restrictions as a model's expectations for a verb and object to appear together in a simple active transitive sentence, and ask whether BERT can make generalizations about selectional restrictions from indirect evidence, following the incomplete selectional network given in Fig. \ref{fig:selectional-restrictions} (a). Indirect evidence plays an important role in human language learning. The role of indirect negative evidence has been the focus of much debate in discussions of innate human learning biases \cite{marcus1993negative, clark2010linguistic}, and indirect evidence has also been shown to play an important role in the learning of novel verbs in both adults and children \cite{perek2017linguistic, yuan2009really, gropen1989learnability}

To assess BERT's ability to leverage indirect negative evidence for verbal selection classes, we fine-tune the model on 12 sentences with verb/object pairings that correspond to the solid lines in Figure \ref{fig:selectional-restrictions} (a). The fine-tuning set (and the test set) consist of simple transitive sentences, following the form ``The [MASK] [Verb1] the [Noun1]." Each novel verb and each novel noun occur twice in the fine-tuning set, meaning that this test assesses the model's few-shot generalization capabilities. The network of verb-noun relations in the 12 fine-tuning sentences implicitly creates two classes of verbs: verbs within a class can be connected with a path through the solid lines. If the model leverages this incomplete evidence to make class-based generalizations, we predict that novel in-class verb/object pairings (which we indicate with dashed lines in the figure) should be more expected than novel out-class verb-object pairings, despite neither having been directly attested in the fine-tuning data.

In order to assess the learning outcome of the model, we follow our psycholinguistic generalization test methodology to derive the probabilities of the verbs in simple active transitive sentences across three testing contexts: In the \textit{attested in-class} condition, we compute the average probability of the verbs in sentences where they are paired with their nouns seen during fine-tuning. This set consisted of 12 sentences, corresponding to the solid lines in Figure \ref{fig:selectional-restrictions} (a). In \textit{unattested in-class} we compute the probability of the verbs when paired with their unattested, but in-class nouns. This set consisted of 6 sentences, corresponding to the dashed lines in Figure \ref{fig:selectional-restrictions} (a). In the \textit{unattested out-class} we compute the probability of the verbs when paired with nouns from the other class. This set consisted of 18 sentences, corresponding to verb-noun combinations that are not connected by lines in Figure \ref{fig:selectional-restrictions} (a).

The results of this experiment can be seen in Figure \ref{fig:selectional-restrictions} (b) and (c). Part (c) shows the average \textit{surprisal} (or negative log probability) of the verbs in the three testing contexts. In (b) we see model `accuracy', or the proportion of times the model assigns lower surprisal to the higher evidence verb/object pairs. For example, for the \textit{attested in-class} vs. \textit{unattested in-class} the y-axis is the proportion of the time the \textit{attested in-class} verbs are given lower surprisal. Results are averaged across all six novel verbs and the proportions are taken accross 200 random model seeds. Our predictions are as follows: For the accuracy test, if the model is able to pick up patterns in the fine-tuning data, we expect the comparison between seen items and unseen items to be greater than the 50\% random baseline. If the model is able to go beyond the patterns in the training data and make class-based generalizations, then we expect the \textit{unattested in-class vs. unattested out-class} comparison, too, to be higher than the baseline.

Examining verb surprisal on the right, we see significant contrasts between each of the conditions (p$<$0.001); crucially, the \textit{unattested in-class} pairings are less surprising (i.e. higher probability) than the \textit{unattested out-class} pairings, despite the model having seen neither pairing during training. This pattern is confirmed with the accuracy scores, where all three contrasts are significantly higher than the 50\% random baseline (p$<$0.001). These results provide strong evidence that BERT is not only sensitive to the minimal amount of data on which it was fine-tuned, but also able to leverage indirect evidence during fine-tuning to make syntactic generalizations, which drive behavior at test time.

\section{Verb Alternation Classes}
\label{Alternations}

\begin{figure*}[t]
\begin{center}
Psycholinguistic Generalization Tests
\includegraphics[width=\textwidth]{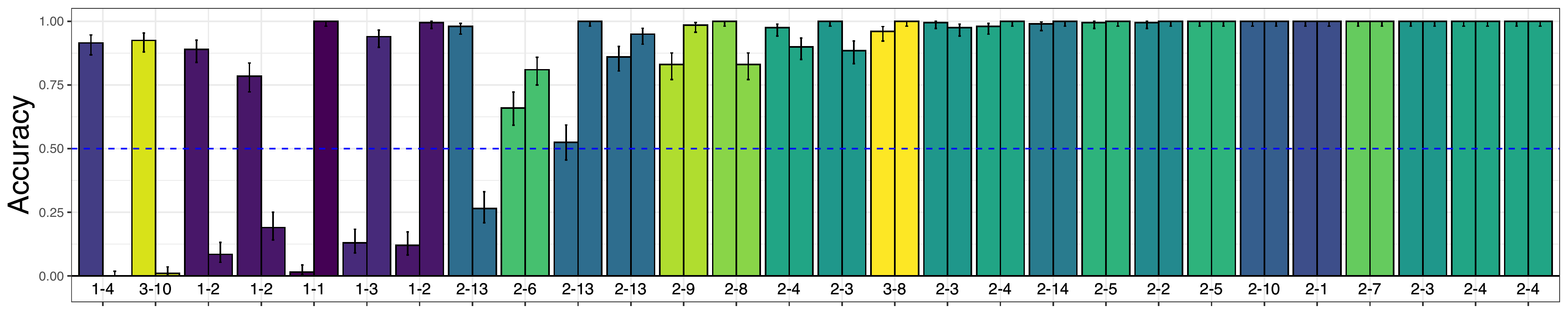}
Key Examples:
\includegraphics[width=0.9\textwidth]{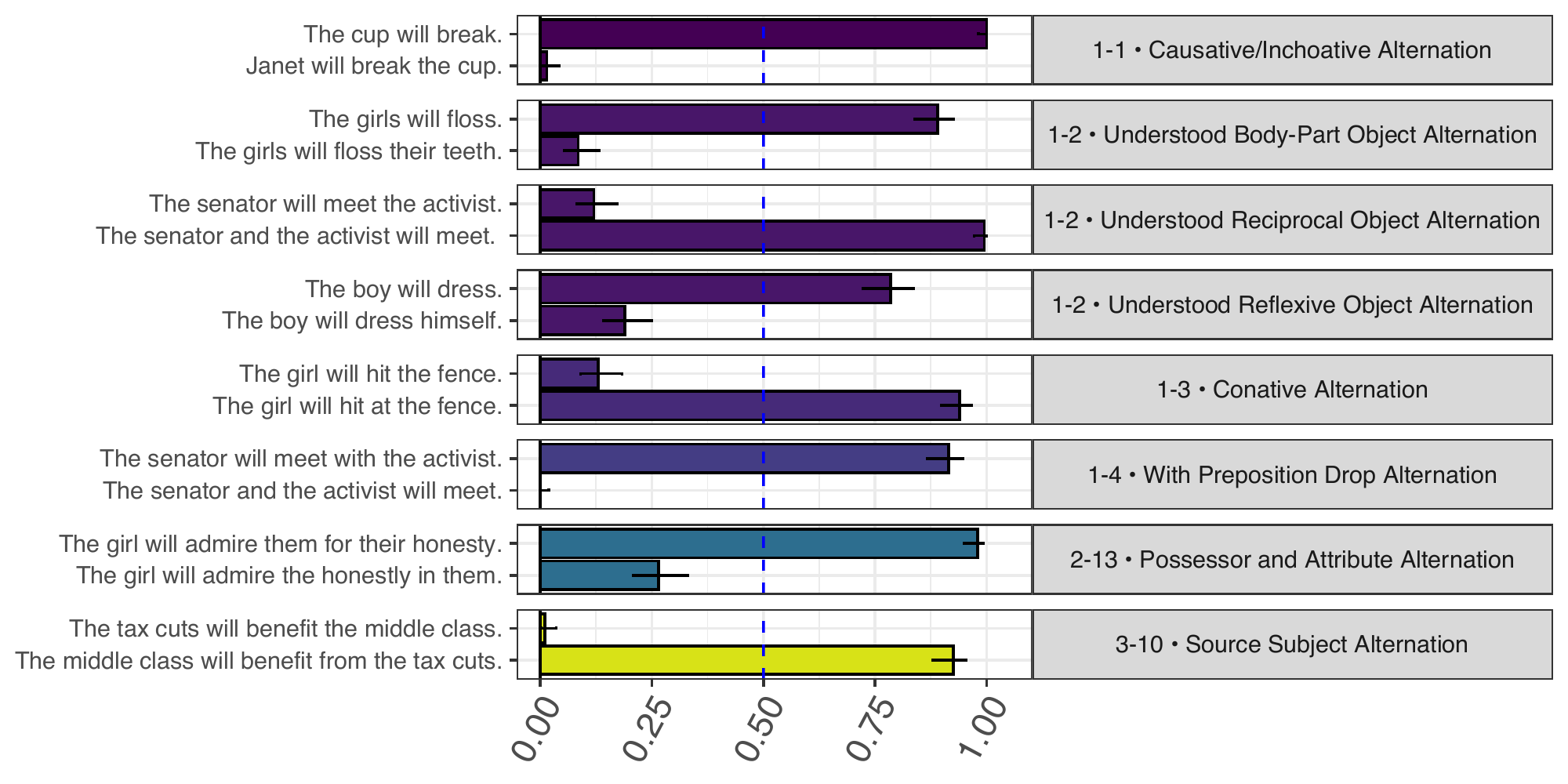}
\end{center}
\caption{Psycholinguistic generalization test accuracy to sister frames by verbal alternation, colored by section and subsection from \cite{levin1993english}. Top figure shows accuracy scores for all alternations tested. Bottom figure shows detailed information for alternations where one frame achieved lower than 50\% accuracy. Error bars show 95\% binomial confidence intervals across 200 random seeds; blue dashed line is the random baseline.}
\label{fig:syntactic_generalization_alternations}
\end{figure*}

English is attested to have at least 83 distinct verbal alternation classes, which were analyzed and categorized in meticulous detail in \citet{levin1993english}. In these experiments we consider all verbal alternation classes for which there are two constant frames and for which Levin provides a list of example verbs as well as a list of ``distractor" verbs---verbs that fit in one frame but not the other---which we require for our embedding classification test paradigm. All of the alternation classes we test come from the first three sections of Levin's `English Verb Classes and Alternations.' To give a brief flavor of the range of English verbal alternations, we give three examples below. 

\ex.Understood Reciprocal Alternation \label{ex:class-1}
\a. The senator will meet the activist.
\b. The senator and the activist will meet.

\ex. Spray/Load Alternation \label{ex:class-2}
\a. The girl will spray the wall with paint.
\b. The girl will spray paint onto the wall.

\ex. Raw Material Subject Alternation \label{ex:class-3}
\a. The girl will make wonderful bread from that flour.
\b. That flour will make wonderful bread.

Verbs like \textit{meet} in Example \ref{ex:class-1} undergo transitivity alternations, where the verb takes a direct object in one frame but not the other. Verbs like \textit{spray} in Example \ref{ex:class-2} involve alternations for transitive verbs that take more than one non-subject argument, and allow for multiple ways of expressing the arguments. Verbs like \textit{make} in Example \ref{ex:class-3} involve ``oblique" subject alternations, where the verb takes one fewer argument in one verbal frame. It is important to note that Levin makes a categorical distinction between these three types of verbal alternation classes and analyzes them each in their own section.

\begin{figure*}
\begin{center}
Embedding Classification Test
\includegraphics[width=\textwidth]{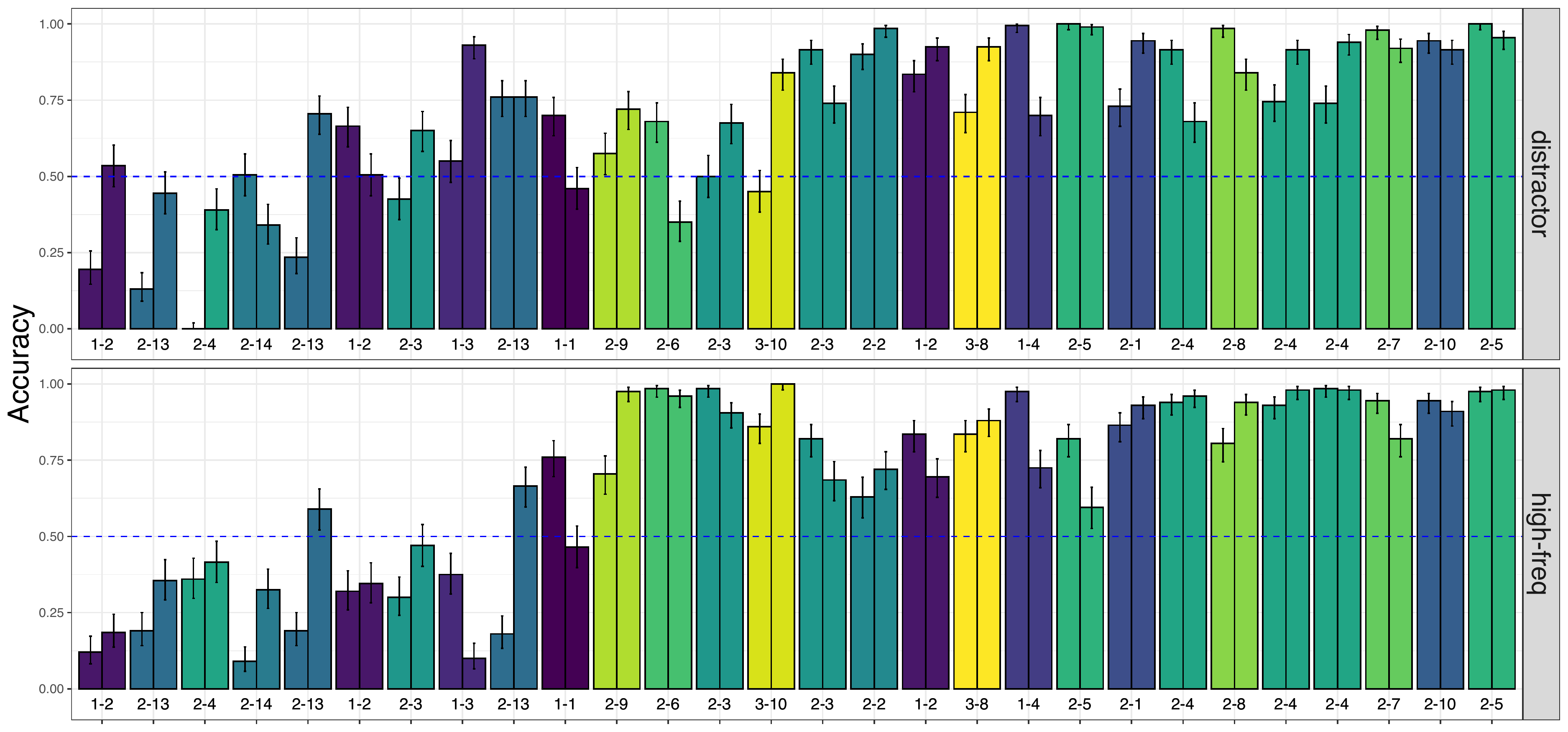}
\end{center}
\caption{Accuracy scores from our embedding classification test, by \textit{out-class} verbs on the different panels. Random baseline is the blue dotted line. Bars are colored by section and subsection from \cite{levin1993english}. Error bars show 95\% binomial confidence intervals across 200 random seeds.}
\label{fig:word_classifier_alternations}
\end{figure*}

For each of the attested alternations, we create one fine-tuning sentence for each frame using the example frames provided by Levin. We replace the attested verb from the example with a novel verb token and mask content words as discussed in Section \ref{sec:methods}.\footnote{Examples of each alternation class and fine-tuning sentences can be found in Appendix A.} We provide tests using both the psycholinguistic generalization and the embedding classification methodology. These are two different ways of probing the generalizations that BERT is able to make, but they result in qualitatively similar results. For our psycholinguistic assessment test, we derive the probability of the novel verb in its alternation-pair frame (this is the \textit{in-class} context), and the mean probability of the verb across all of the other verbal frames that do not form one of our alternation classes with the training frame (these are the \textit{out-class} contexts). For our embedding classification test, we train two classifiers for each frame: The first predicts between attested verbs that follow one of the frame's alternations provided by Levin, and a set of out-class \textit{distractor} verbs that can appear in one of the frames but not the other, also provided by Levin. The second predicts between the attested verbs and an out-class set of the 150 most frequent verbs from the Corpus of Contemporary American English (COCA) \cite{davies2015coca}, pruned of auxiliary and modal verbs and verbs that already appear in Levin's lists. For each verbal alternation, we run two classification tests, one for each frame in the alternation.

\subsection{Psycholinguistic Assessment Results}

The results from our psycholinguistic assessment test can be seen in Figure \ref{fig:syntactic_generalization_alternations}. On the top row we show mean accuracy scores across 200 random seeds for all of the alternations tested. On the bottom panel we zoom in on a few key examples, specifically instances where the model performs below the 50\% baseline on one of the training frames. Here, we have flipped the axes for readability. For each alternation tested, our charts include two bars, which correspond to the two separate training frames. These training frames are labeled in the bottom figure, with the label corresponding to the type of sentence that we fine-tune the model on. If the model shows high accuracy scores on both bars, it means it has learned the bi-directionality of the alternation. If it shows high accuracy scores in only one training frame, however, it means that it has only learned to generalize from that frame to its sister. Across all our figures, alternations are colored and labeled by the section and first-level subsection of \citet{levin1993english} (e.g. \texttt{1-4} means Section 1 Subsection 4, etc.). Error bars are 95\% binomial confidence intervals across the 200 random seeds. To see a full-breakdown of all alternations and training frames tested see Appendix \ref{sec:psycholinguistic-results}.

In terms of top-level performance, BERT performs quite well. Across all alternation classes, the model achieves 82\% accuracy, which is significantly higher than the  50\% random baseline (p$<$0.001) and for about half the alternation tests, BERT achieves accuracy scores that are at, or near 100\%. Note that the model's performance at these tests generally corresponds to the top-level subsection of \cite{levin1993english}, with generally higher scores from Sections 2 and 3, and lower scores from Section 1 (darker blue and purple bars), which correspond to alternations that involve a change in transitivity. Another observation is that when the model does fail, it does so for only one of the two frames. For all cases where the model performs below baseline on one of the training frames, it performs at, or above 75\% accuracy on the other frame.

Zooming in on the cases where the model fails to generalize, we see a robust pattern: Almost all cases where model accuracy scores are below 25\% are for transitive alternation frames in which the model is being fine-tuned on a single example with a direct object and asked to generalize to cases where the direct object is absent. For example, with the \textit{Understood Reflexive Object Alternation} BERT was $\sim$25\% accurate when fine-tuned with the frame of the example ``The boy will [\textit{nonce}] himself" but $\sim$75\% accurate when fine-tuned with the frame of ``The boy will [\textit{nonce}].'' At a high level, this means that given a single instances of a verb without an object, models expect that it will occur with a direct object, at least more-so than with oblique or prepositional objects (the various \textit{out-class} frames). However, when given a single instance of a transitive verb, models do not expect it to occur intransitively. The fact that tokens seen only a few times are generally expected to be able to take direct objects suggests a transitivity learning bias in the model. Such a bias would align with recent work assessing few-shot learning of syntactic categories, specifically \citet{jumelet2019analysing}, who hypothesize that models learn default category for number and gender, and \citet{wilcox2020fewshot}, who provide data from few-shot learning tests that is consistent with the hypotheses in \citet{jumelet2019analysing}. Interestingly, the results form \citet{wilcox2020fewshot} also suggest that the models tested learn a default \textit{transitive} category for verbs, although they test Recurrent Neural Network models, not transformers, so more careful cross model comparisons are needed.

\subsection{Classificaiton Assessment Results}

The results from our classification assessment test can be seen in Figure \ref{fig:word_classifier_alternations}. Accuracy scores are on the y-axis and verbal alternation classes are on the x-axis, with the results from the \textit{distractor} out-class on the top panel and the \textit{high-frequency} out-class on the bottom panel. Across all verbal alternations and out-class groups tested, BERT achieves an average accuracy of 69\%, which is significantly higher than the 50\% baseline (p$<$0.001), and does not perform significantly better or worse on either the distractor or high-frequency out-classes (p=0.6). As before, the model performs generally worse on alternations from Section 1 of \cite{levin1993english}, although BERT's performance on the classification assessment test is much more varied than its performance on the psycholinguistic assessment tests. That being said, the scores are correlated (rank performance $cor=0.49, p<0.001$; raw accuracy scores $cor=0.17, p=0.08$). 

\section{Conclusion}

We used a novel word learning paradigm, inspired by classic studies from psycholinguistics, to assess BERT's syntactic generalization behavior on two novel phenomena: English verb class alternations and verb/object selectional restrictions.  In both cases we address the issue of single and few-shot learning by fine-tuning the model on just one or two positive examples, finding that BERT makes some generalizations about a novel token based on minimal experience, and that these generalizations drive robust behavior during test time. This novel word learning paradigm can continue to be explored in later work through the use of large databases such as VerbNet \cite{schuler2005verbnet}, which builds on Levin's verb documentations by providing a larger database of verb alternations and sectional restrictions that can be turned into train and test sentences for BERT without hand-crafting.

For verbal/object selectional restrictions, we find that BERT leverages indirect evidence to expect unattested but plausible verb/noun pairings more than unattested but implausible pairings. These results provide evidence for the view that the model is able to attend not just to patterns overtly realized in the data (direct evidence) but also implicit relationships between tokens (indirect evidence). The ability to use indirect evidence, specifically indirect \textit{negative} evidence, is a hallmark of human language learning, and these results indicate that models are capable of similar behavior in a simple novel word learning paradigm.

For verbal alternations, we find that when fine-tuned on a single frame, BERT routinely expects the verb to occur in its sister frame with a higher likelihood than in unrelated verbal frames. Interestingly, this behavior is consistently blocked when the model is asked to generalize from a frame that involves an object to a frame where the object is lacking. This behavior is consistent with a general bias towards transitivity in the model, and suggests an exciting direction for further study. Whether such a general bias exists, whether it is restricted to settings with limited evidence, and whether it changes as verbs appear more frequently in the fine-tuning or training data is a question for future research. Another question for future research is whether a multilingual BERT would have the same success on alternation tests in other languages, and if if would exhibit the same biases that we see for English.

\section*{Acknowledgments}

We gratefully acknowledge support from the MIT–IBM AI Research Lab and a Google Faculty Research Award.

\bibliographystyle{acl_natbib}
\bibliography{anthology,emnlp2020}

\appendix

\section{Supplemental Alternation Material}
\label{sec:supplemental_alternations}

Each subsection contains Levin's example of an alternation, followed by training data for BERT that exemplifies the alternation with a novel verb token: [Vn]. A ``distractor'' example from Levin of a verb that does not follow the alternation is also given.

\subsection{Causative/Inchoative}

Janet broke/forfeited the cup. The cup broke/*forfeited.\\
The [MASK] will [V1.1] the [MASK]. The [MASK] will [V1.2].

\subsection{Understood Body-Part Object}

I flossed/bumped my teeth. I flossed/*bumped.\\
The [MASK] will [V2.1]. The [MASK] will [V2.2] their [MASK].

\subsection{Understood Reflexive Object}

Jill dressed/groomed herself hurriedly. Jill dressed/*groomed hurriedly.\\
The [MASK] will [V3.1]. The [MASK] will [V3.2] themself.

\subsection{Understood Reciprocal Object}

Anne met/*agreed Cathy. Anne and Cathy met/agreed.\\
The [MASK] will [V4.1] the [MASK]. The [MASK] and the [MASK] will [V4.2].

\subsection{Conative}

Paula hit/bonked the fence. Paula hit/*bonked at the fence.\\
The [MASK] will [V5.1] the [MASK]. The [MASK] will [V5.2] at the [MASK].

\subsection{\textit{with} Preposition Drop}

Jill met/*kissed with Sarah. Jill met/kissed Sarah.\\
The [MASK] will [V6.1] with the [MASK]. The [MASK] will [V6.2] the [MASK].

\subsection{Dative}

Bill sold/surrendered a car to Tom. Bill sold/*surrendered Tom a car.\\
The [MASK] will [V7.1] a [MASK] to the [MASK]. The [MASK] will [V7.2] the [MASK] a [MASK].

\subsection{Benefactive}

Martha carved/confiscated a toy for the baby. Martha carved/*confiscated the baby a toy.\\
The [MASK] will [V8.1] a [MASK] for the [MASK]. The [MASK] will [V8.2] the [MASK] a [MASK].

\subsection{Spray/Load}

Jack sprayed/dumped paint on the wall. Jack sprayed/*dumped the wall with paint.\\
The [MASK] will [V9.1] the [MASK] onto the [MASK]. The [MASK] will [V9.2] the [MASK] with the [MASK].

\subsection{Clear (transitive)}

Henry cleared/extracted dishes from the table. Henry cleared/*extracted the table of dishes.\\
The [MASK] will [V10.1] the [MASK] from the [MASK]. The [MASK] will [V10.2] the [MASK] of the [MASK].

\subsection{Swarm}

Bees are swarming/clustering in the garden. The garden is swarming/*clustering with bees.\\
The [MASK] will [V11.1] in the [MASK]. The [MASK] will [V11.2] with the [MASK].

\subsection{Material/Product (transitive)}

Martha carved/*turned a toy out of the piece of wood. Martha carved/turned the piece of wood into a toy.\\
The [MASK] will [V12.1] a [MASK] out of the [MASK]. The [MASK] will [V12.2] the [MASK] into a [MASK].

\subsection{Material/Product (intransitive)}

That acorn will grow/turn into an oak tree. An oak tree will grow/*turn from that acorn.\\
That [MASK] will [V13.1] into an [MASK]. An [MASK] will [V13.2] from that [MASK].

\subsection{Total Transformation (transitive)}

The witch turned/compiled him into a frog. The witch turned/*compiled him from a prince into a frog.\\
The [MASK] will [V14.1] the [MASK] into a [MASK]. The [MASK] will [V14.2] the [MASK] from a [MASK] into a [MASK].

\subsection{Total Transformation (intransitive)}

He turned/grew into a frog. He turned/*grew from a prince into a frog.\\
The [MASK] will [V15.1] into a [MASK]. The [MASK] will [V15.2] from a [MASK] into a [MASK].

\subsection{\textit{apart} Reciprocal (transitive)}

I broke/disconnected the twig off (of) the branch. I broke/*disconnected the twig and the branch apart.\\
The [MASK] will [V16.1] the [MASK] off of the [MASK]. The [MASK] will [V16.2] the [MASK] and the [MASK] apart.

\subsection{\textit{apart} Reciprocal (intransitive)}

The twig broke/disconnected off (of) the branch. The twig and the branch broke/*disconnected apart.\\
The [MASK] will [V17.1] off of the [MASK]. The [MASK] and the [MASK] will [V17.2] apart.

\subsection{Fulfilling}

The judge presented/offered a prize to the winner. The judge presented/*offered the winner with a prize.\\
The [MASK] will [V18.1] a [MASK] to the [MASK]. The [MASK] will [V18.2] the [MASK] with a [MASK].\\

\subsection{Image Impression}

The jeweller inscribed/transcribed the name on the ring. The jeweller inscribed/*transcribed the ring with the name.\\
The [MASK] will [V19.1] the [MASK] on the [MASK]. The [MASK] will [V19.2] the [MASK] with the [MASK].

\subsection{\textit{with}/\textit{against}}

Brian hit/threw the stick against the fence. Brian hit/*threw the fence with the stick.\\
The [MASK] will [V20.1] the [MASK] against the [MASK]. The [MASK] will [V20.2] the [MASK] with the [MASK].

\subsection{\textit{through}/\textit{with}}

Alison pierced/*hit the needle through the cloth. Alison pierced/hit the cloth with a needle.\\
The [MASK] will [V21.1] the [MASK] through the [MASK]. The [MASK] will [V21.2] the [MASK] with a [MASK].

\subsection{\textit{blame}}

Mira blamed/*hated the accident on Terry. Mira blamed/hated Terry for the accident.\\
The [MASK] will [V22.1] the [MASK] on the [MASK]. The [MASK] will [V22.2] the [MASK] for the [MASK].

\subsection{Possessor Object}

They praised/detected the volunteers' dedication. They praised/*detected the volunteers for their dedication.\\
The [MASK] will [V23.1] their [MASK]. The [MASK] will [V23.2] them for their [MASK].

\subsection{Attribute Object}

I admired/praised his honesty. I admired/*praised the honesty in him.\\
The [MASK] will [V24.1] their [MASK]. The [MASK] will [V24.2] the [MASK] in them.

\subsection{Possessor and Attribute}

I admired/*detected him for his honesty. I admired/detected the honesty in him.\\
The [MASK] will [V25.1] them for their [MASK]. The [MASK] will [V25.2] the [MASK] in them.

\subsection{\textit{as}}

The president appointed/declared Smith press secretary. The president appointed/*declared Smith as press secretary.\\
The [MASK] will [V26.1] the [MASK] the [MASK]. The [MASK] will [V26.2] the [MASK] as the [MASK].

\subsection{Raw Material Subject}

She baked/invented wonderful bread from that whole wheat flour. That whole wheat flour bakes/*invents wonderful bread.\\
The [MASK] will [V27.1] the [MASK] from that [MASK]. That [MASK] will [V27.2] the [MASK].

\subsection{Source Subject}

The middle class will benefit/gain from the new tax laws. The new tax laws will benefit/*gain the middle class.\\
The [MASK] will [V28.1] from the [MASK]. The [MASK] will [V28.2] the [MASK].

\section{Model Details}

\subsection{BERT tuning}
BERT version = bert-large-uncased from \url{https://github.com/huggingface/transformers}, Optimizer = Adam \cite{kingma2015adam}, learning rate = 1e-3, batch size = full training set size (each training sentence is a separate datum and is enclosed by a start and end token), epochs = 10.

\subsection{Linear Classifier}
Architecture = linear layer with an input size the same as that of a BERT embedding and an output size of 2, optimizer = Adam, learning rate = 1e-1, batch size = full training set, epochs = 20, loss = Cross Entropy; trained to label a datum as in-class or out-class with labels of 1 and 0, respectively.

\section{Psycholinguistic Generalization Test: Full Breakdown} \label{sec:psycholinguistic-results}

\begin{figure*}[t]
\begin{center}
Psycholinguistic Generalization Tests: Full Information
\includegraphics[width=\textwidth]{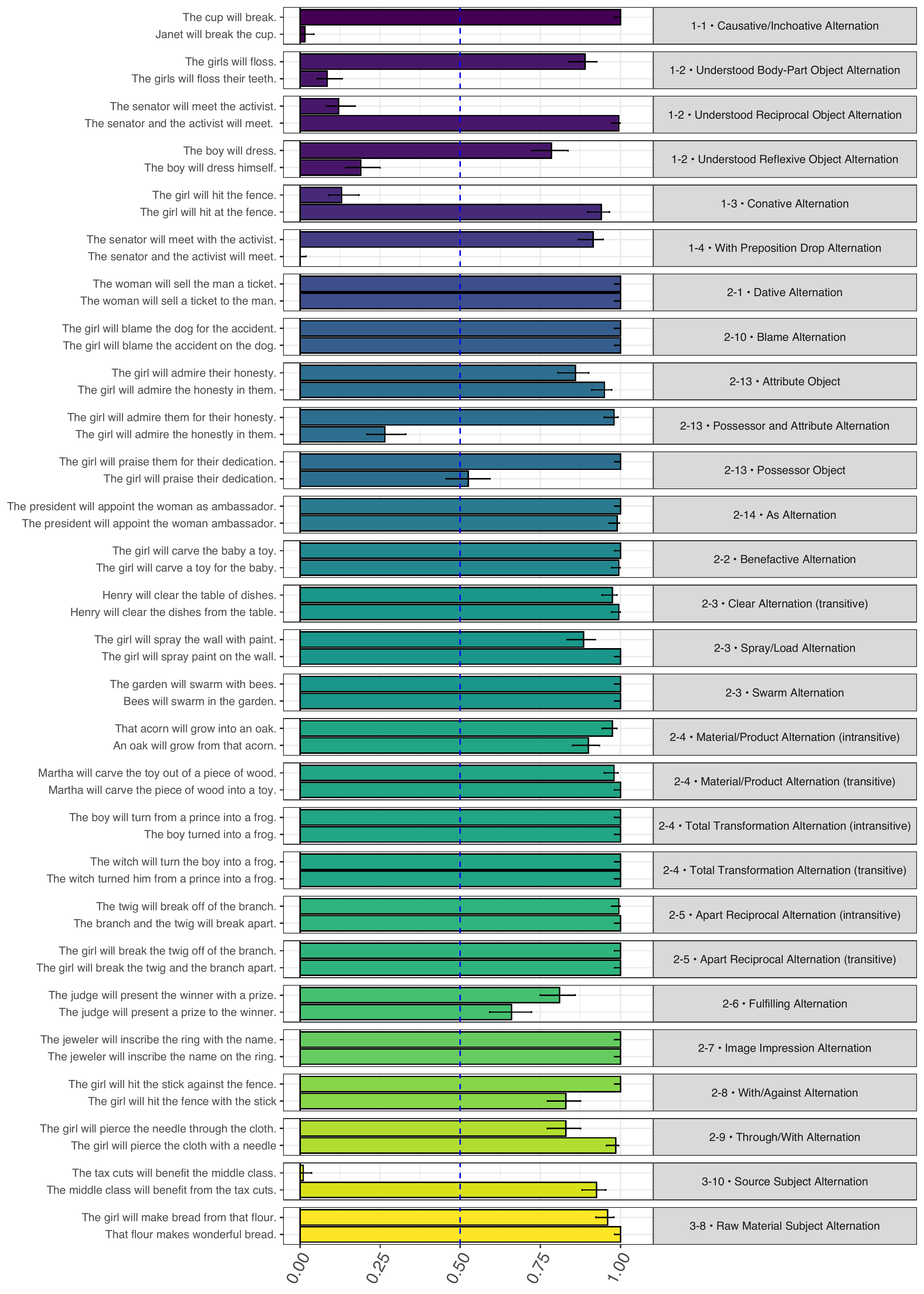}
\end{center}
\caption{Psycholinguistic generalization test accuracy to sister frames by verbal alternation, colored by section and subsection from \cite{levin1993english}. Error bars show 95\% binomial confidence intervals across 200 random seeds; blue dashed line is the random baseline.}
\label{fig:syntactic_generalization_alternations_full}
\end{figure*}

\end{document}